%% file: preprint.tex
\documentclass{article}
\usepackage[preprint]{neurips_2026}
\usepackage{amsmath}
\usepackage{multirow}
\usepackage{float} 
\usepackage[utf8]{inputenc}
\usepackage[T1]{fontenc}   
\usepackage{hyperref}      
\usepackage{url}           
\usepackage{booktabs}      
\usepackage{amsfonts}      
\usepackage{nicefrac}      
\usepackage{microtype}     
\usepackage{xcolor}        
\usepackage{graphicx}

\title{Quantum Advantage in Multi Agent Reinforcement Learning}

\author{
  Simranjeet Singh Dahia, Claudia Szabo \\[0.3em]
  Adelaide University, Australia \\[0.3em]
  \small \texttt{\{simranjeetsingh.dahia, claudia.szabo\}@adelaide.edu.au}
}

\begin{document}

\maketitle

\begin{abstract}
We present an empirical evaluation of quantum entanglement in agent coordination within quantum multi agent reinforcement learning (QMARL). While QMARL has attracted growing interest recently, most prior work evaluates quantum policies without provable baselines, making it impossible to rigorously distinguish quantum advantage from algorithmic coincidence. We address this directly by evaluating a decentralized QMARL framework with variational quantum circuit (VQC) actors with shared entangled states.  In the CHSH game, which has a mathematically proven classical performance ceiling of 0.75 win rate, we show that \textit{entangled} QMARL agents approach the Tsirelson limit of 0.854, providing clear evidence of their quantum advantage. We show that unentangled quantum circuits match the classical baseline, confirming that entanglement and not the quantum circuit itself is the active coordination mechanism. We also explore the effect of specific entanglement structures, as some Bell states enable coordination gains while others actively harm performance. On cooperative navigation (CoopNav), QMARL without entanglement achieves $\sim2\times$ improvement in success rate over classical MAA2C ($\sim$0.85 versus $\sim$0.40), with the hybrid configuration, quantum actor paired with a classical centralised critic, outperforming both fully classical and fully quantum solutions. We present our experimental analysis and discuss future work.
\end{abstract}

\section{Introduction}
Multi agent reinforcement learning (MARL) has emerged as the leading computational paradigm for modelling sequential decision making in multi agent systems \cite{zhang_multi-agent_2021, hernandez2019surveyppr}. By treating each component as an independent learning agent, MARL provides a mechanism to capture the non-linear interdependencies and emergent dynamics that characterise complex systems \cite{oroojlooyjadid_review_2021} and  centralised training with decentralized execution (CTDE) paradigms, and approaches solving challenges such as cooperative policy gradient \cite{lowe_multi-agent_2020} and value decomposition \cite{sunehag_value-decomposition_2017}, have produced appealing results across games, robotic coordination and network optimisation \cite{kraemer_multi-agent_2016}. However, a consequence of decentralized partially observable Markov Decision Process (Dec-POMDP), classical MARL agents acting on local observations may converge to locally optimal but globally suboptimal policies, as independently acting agents cannot access each other's observations at execution time \cite{wen_multi-agent_2022}. Overcoming this requires either explicit communication (incurring bandwidth cost and scalability bottlenecks) \cite{marl_learning_to_communicate_2016}, centralised coordination (sacrificing privacy, autonomy and robustness) \cite{bernstein_complexity_2002} or communication-free coordination mechanisms that implicitly align agent policies \cite{wen_multi-agent_2022}.

Quantum mechanics offers a compelling candidate for a training and coordination mechanism where agents are aware of their training and states by design, namely, \textit{entanglement}. When agents share a pre-prepared entangled quantum state, their local measurement outcomes are intrinsically correlated in ways that have no classical formulation \cite{brunner2014bellnonlocality}. Importantly, in decentralized MARL scenarios, by design these correlations are established before execution and exploited locally without any classical communication between agents at runtime.
This property has been extensively studied in quantum information theory and quantum game theory (QGT) ~\cite{Eisert1999QuantumGames}, where quantum strategies have been shown to break classical performance limits in coordination tasks \cite{brunner2014bellnonlocality, Eisert1999QuantumGames}. Our central hypothesis is that this same mechanism can serve as an implicit coordination mechanism for MARL agents acting in decentralized settings, enabling modelling of behaviors that are otherwise inaccessible with classical independent agents.

\begin{figure}[htbp]
    \centering
    \includegraphics[width=0.91\linewidth]{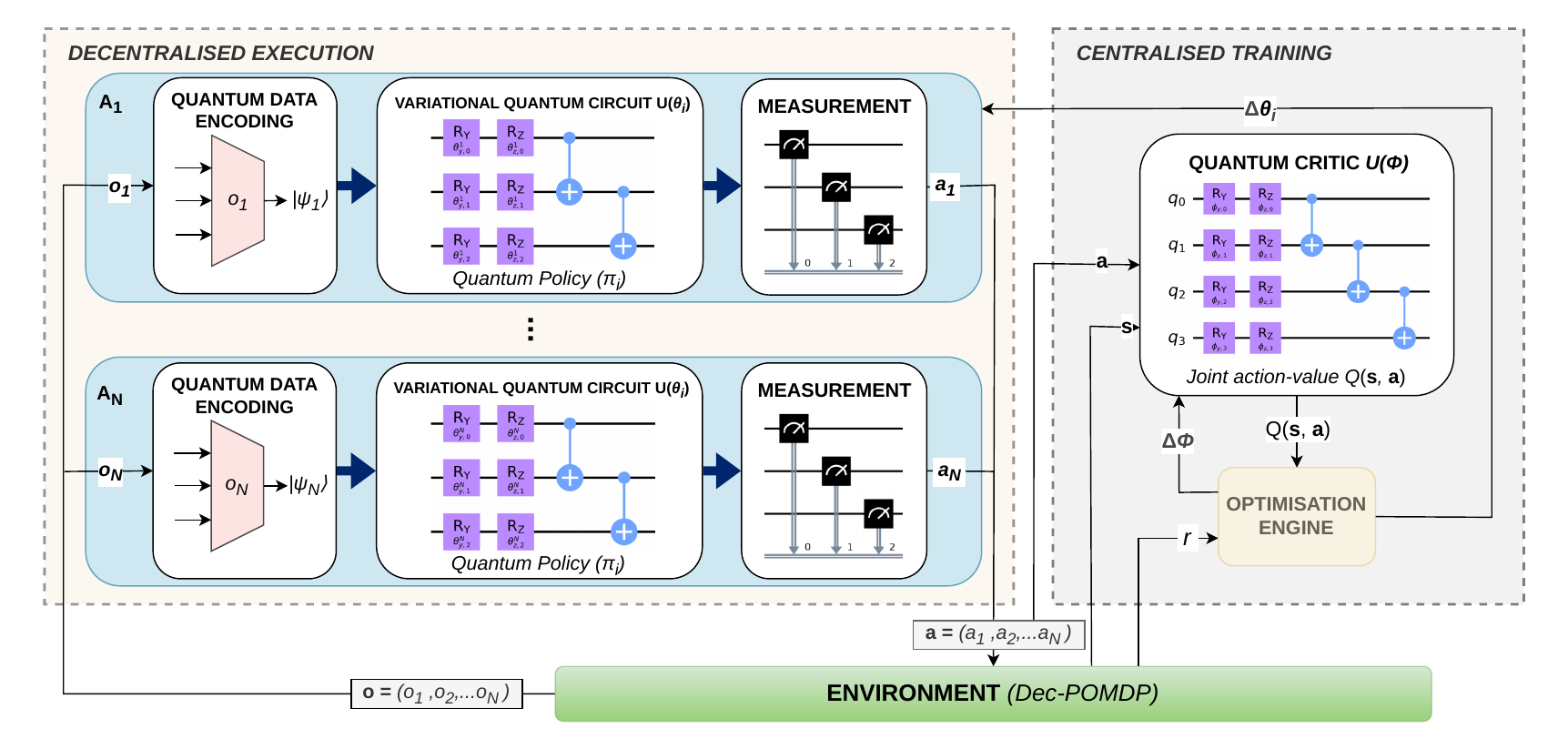}
    \caption{QMARL under CTDE: Agents execute decentralised 
    quantum policies independently; a centralised critic trains all agents using global state and joint actions.
    }

    \label{fig:qmarl_framework} 
\end{figure}

In decentralized multi agent settings with emergent dynamics, the bottleneck is rarely computation speed; it is the quality and structure of agent coordination \cite{bernstein_complexity_2002}. 
Independent agents, no matter how powerful, converge to suboptimal equilibrium because their policies lack access to global coordination information ~\cite{wen_multi-agent_2022}.
We argue that quantum computing does not merely accelerate existing MARL algorithms. Using quantum mechanisms has the potential to enable a different class of coordination strategies that are provably beyond the reach of classical decentralized independent agents \cite{brunner2014bellnonlocality, clauser_proposed_1969}.

Quantum reinforcement learning (QRL) and quantum MARL (QMARL) have attracted growing attention ~\cite{yun_quantum_2022, chen_qmarl_2024, jerbi2021_parametrized_quantum_rl}, with hybrid quantum-classical approaches using
variational quantum circuits (VQCs) as parameterised policy
networks ~\cite{skolik_quantum_2022}. However, existing work predominantly substitutes classical network components with quantum analogues, and focuses on  whether VQCs can replace classical neural networks. In contrast, we focus on understanding the impact of quantum phenomena such as entanglement and non-local correlations on exploration-exploitation. In addition, most QMARL results
are evaluated on toy benchmarks without provably tight classical baselines, making it impossible to rigorously claim quantum advantage ~\cite{qmarl_vqc_design_2022, kesiku_how_2026}. In this work, we focus on the rigorous analysis of the benefits of entanglement.  We select a problem with proven classical performance limit, where no classical decentralized strategy can exceed a known upper bound and define quantum advantage as whether a  QMARL agent, exploiting entanglement, exceeds this proven bound. 
The \textbf{contribution} of our work is threefold: (i) We establish that quantum advantage in decentralized multi agent systems can arise from entanglement-based coordination. (ii) We demonstrate that decentralized quantum agents sharing
entangled states exceed the classical performance
ceiling and that that entanglement structure is
critical, with some entanglement variants actively harming coordination. (iii) We show that a  hybrid quantum-classical CTDE outperforms both fully classical and fully quantum CTDE on cooperative navigation benchmarks.

\section{Related Work}
Classical MARL formalises cooperative and competitive multi agent problems as Dec-POMDPs ~\cite{bernstein_complexity_2002}, where decentralized agents acting on partial observations must learn collectively optimal policies without runtime communication. This is optimized through Centralised training with decentralized execution (CTDE) ~\cite{lowe_multi-agent_2020} which allows a global critic during training while preserving decentralized execution.
Our work adopts CTDE as the training framework and asks whether replacing classical actors with variational quantum circuits changes what coordination is achievable.

We extend VQC-based policies to the multi agent settings under CTDE and ask whether entanglement, rather than quantum circuits alone, enables a more sophisticated coordination mechanism. The most directly related work ~\cite{eQMARL} proposes an entanglement assisted CTDE framework where a shared quantum state couples decentralized agents through a quantum critic. eQMARL demonstrates coordination gains in small cooperative settings and is the closest prior work to ours in motivation. However, it evaluates exclusively on arbitrary MARL benchmarks without provably tight classical baselines, making it impossible to attribute improvements specifically to entanglement rather than to model capacity or algorithmic differences. Our work directly addresses this gap by grounding evaluation in a problem with a mathematically proven classical ceiling.

Another work closely related to our CHSH experiments ~\cite{gardiner_learning_2026} demonstrates quantum advantage in nonlocal games by training agents conditioned on the outputs of a centralised quantum coordinator. Their approach achieves strong results but the coordinator acts as a shared model at runtime, making the architecture quasi-centralised rather than a truly decentralized multi agent setting. The authors explicitly acknowledge this limitation, noting that their framework provides coordination advantages at training time. Our setup is strictly decentralized; each agent is an independent VQC that measures only its own share of a pre-prepared entangled state, with no shared model at runtime. This represents a harder and more realistic problem setting, requiring agents to independently learn to exploit entanglement without centralised coordination.

Quantum strategies in two-player games were studied by Eisert et al. ~\cite{Eisert1999QuantumGames} and Meyer~\cite{meyer1999_quantum_strategies}, showing that entangled players can access payoff regions unavailable to classical strategies. Brunner et al. ~\cite{brunner2014bellnonlocality} provide the definitive treatment of Bell nonlocality, of which the CHSH inequality ~\cite{clauser_proposed_1969} is the prime example. The Tsirelson bound ~\cite{cirelson_quantum_1980} establishes the possible quantum maximum at $\cos^2(\pi/8) \approx 0.854$. These results form the theoretical foundation for our use of CHSH as a calibration benchmark: it provides a problem where classical performance is provably bounded, the quantum target is known, and any consistently higher win rate constitutes unambiguous evidence of quantum advantage. To our knowledge, we are the first to use CHSH as a rigorous calibration benchmark within a full MARL training loop with truly decentralized agents.

\section{Method}
\label{sec:methods}
Our work focuses on understanding the benefits of using a QMARL versus a MARL framework, and in particular determining the parts of the framework that benefit from a quantum formulation and of integrating quantum entanglement in the MARL pipeline. In our experimental analysis, we consider three environments with increasing complexity, CHSH, with a provable classical bound, and CoinGame and CoopNav, both standard MARL benchmark environments. We focus on cooperative multi agent systems in which agents must work together in a shared environment without communication, and may optionally exploit shared quantum entanglement as a coordination resource. This is formalised as decentralized partially observable Markov decision process (Dec-POMDP) ~\cite{bernstein_complexity_2002}.

This section describes the general learning framework, quantum policy architecture and the training algorithm applied across all experiments. Figure ~\ref{fig:qmarl_framework} illustrates the general QMARL framework under CTDE adopted in this work.

We consider cooperative multi agent systems with $\geq 2$ agents interacting in a shared environment. At each time step $t$, the environment is in some global state $s^t$ and each agent $i$ receives a local observation $o_i^t$ from the environment. Using only this local observation, the agent $i$ selects an action $a_i^t$ according to its policy $\pi_i$. All agents act simultaneously, producing a joint action $\textbf{a}^t = (a_1^t, ..., a_N^t)$. Then the environment transitions to a new updated state and all agents receive a shared reward $r^t = r(s^t, \textbf{a}^t)$.

Thus, the goal for the agents is to find policies $\theta_1, ..., \theta_N$ that maximise the team's expected total reward over an episode of length $T$, as:
\begin{equation}
    J(\theta_1, \ldots, \theta_N)
    = \mathbb{E}\!\left[\sum_{t=0}^{T}
      \gamma^t r^t\right]
    \label{eq:objective}
\end{equation}
where $\gamma \in (0,1]$ is the discount factor for future rewards.

At execution time, each agent acts using only its local observation $o_i^t$; no agent has access to another agent's observation, action or internal state during execution.
This is the standard decentralized execution constraint implemented in all our experiments,  making cooperative MARL difficult: \textit{classical agents acting independently on local information converge to suboptimal collective behaviour, a problem that is provably NEXP-complete for $N \geq 2$ agents} ~\cite{bernstein_complexity_2002}.

In contrast, in the quantum variants of our experiments agents  share a pre-prepared entangled quantum state $|\Psi\rangle$  before an episode starts. There are one or more qubits per agent from this entangled state. Thus while each agent applies local quantum operations only to its qubit(s) (private observations) and measures it privately, the outcome is correlated with the outcomes of other agent's measurements, even though no explicit classical information sharing happens between the agents. This is the key property of quantum entanglement ~\cite{brunner2014bellnonlocality}. 
In our experiments we  enforce identical strict no-communication constraints at execution time and consider quantum, entangled and classical implementations. This ensures that any performance gap between quantum and classical agents cannot be attributed to information exchange but only to entanglement.  We use Centralised Training, decentralized Execution (CTDE) as our learning framework, MAA2C ~\cite{mnih_asynchronous_2016} as the training algorithm, with specific considerations for the CHSH game, as discussed in detail in Appendix ~\ref{appendix:learning_framework}.

This work was supported with supercomputing resources provided by the Phoenix HPC service at Adelaide University.

\subsection{Policy Architectures}
\label{sec:policy_arch}
\paragraph{Classical:} The classical actor for agent $i$ is a feedforward neural network mapping local observation $o_i$ to a softmax distribution over discrete actions. The centralised critic is a separate feedforward network mapping the global state to a scalar value estimate $V(s)$. Network architectures are environment-specific and detailed in the respective experiment sections. For classical policies, gradients are computed by standard backpropagation through the neural network.

\paragraph{Quantum:} The quantum actor replaces the classical feedforward network with a variational quantum circuit (VQC). The design follows three stages: state preparation, parameterised local operations and measurement.

\paragraph{Stage 1: Entangled state preparation:} An $N$-qubit entangled state is prepared at the start of each episode. Each agent holds at least one qubit. Supported states include:
\begin{align}
    |\Phi+\rangle &= 
    \tfrac{1}{\sqrt{2}}(|00\rangle + |11\rangle)
    \quad (N{=}2,\text{ symmetric})
    \label{eq:bell}\\
    |\text{GHZ}_N\rangle &= 
    \tfrac{1}{\sqrt{2}}
    (|0\rangle^{\otimes N} + |1\rangle^{\otimes N})
    \quad (N{>}2)
    \label{eq:ghz}
\end{align}
The product state (no entanglement) is included 
as a control in all experiments.

\paragraph{Stage 2: Parameterised local operations:} Each agent applies a sequence of learned single-qubit rotation gates conditioned on its private observation. For depth $d$:
\begin{equation}
    U_i(o_i;\theta_i) 
    = \prod_{\ell=1}^{d} 
      R_Y\!\left(2\theta_i^{(\ell,Y)}[o_i]\right)
      R_Z\!\left(2\theta_i^{(\ell,Z)}[o_i]\right)
    \label{eq:vqc}
\end{equation}
where $\theta_i^{(\ell,Y)}, \theta_i^{(\ell,Z)} \in \mathbb{R}$ are learnable parameters. Gates are strictly local: each agent operates only on its own qubit(s).

\paragraph{Stage 3: Measurement and action decoding:} Qubits are measured in the computational basis. The joint probability table $P(\mathbf{a} \mid \mathbf{o};\theta)$ is computed from the statevector via Born's rule:
\begin{equation}
    P(\mathbf{a} \mid \mathbf{o};\theta) 
    = \left|\langle \mathbf{a} \mid 
      U(\mathbf{o};\theta)\,|\Psi\rangle\right|^2
    \label{eq:born}
\end{equation}
Measurement outcomes are passed through a classical linear layer to produce action probabilities (softmax over discrete actions) in all environments tested here.

CHSH experiments were simulated using Qiskit Aer 0.17.2 \cite{qiskit_2019} (exact, noiseless statevector), calculating analytical gradients explicitly via the parameter-shift rule ($\delta = \pi/4$, requiring two forward passes per parameter) \cite{schuld_evaluating_2019}. The CoinGame and CoopNav environments utilised TensorFlow Quantum 0.7.2 \cite{broughton_tensorflow_nodate} and Cirq \cite{cirq_quantum} for simulating quantum circuits. We integrated these quantum circuits as trainable Keras layers \cite{keras_deep_learning}, leveraging TensorFlow's automatic differentiation to efficiently compute gradients through the quantum layers.

For the CHSH experiment, quantum gradients are computed using the parameter-shift rule:
\begin{equation}
    \frac{\partial}{\partial \theta} 
    \log P(a \mid o;\theta)
    \approx 
    \frac{
        P(a; \theta{+}\delta) 
        {-} P(a; \theta{-}\delta)
    }{
        2\,P(a;\theta)\,\sin(\delta)
    },
    \quad \delta = \frac{\pi}{4}
    \label{eq:param_shift}
\end{equation}

For CoinGame and CoopNav setups, the VQC is implemented as a \texttt{tfq.layers.PQC} layer within a Keras model, allowing end-to-end gradient computation via TensorFlow's automatic differentiation. No explicit parameter-shift computation is required.

\section{Experimental Analysis}
\label{sec:environmental analysis}

Our experimental analysis aims to show the benefits of quantum entanglement in capturing implicit coordination between decentralized agents. In this section, we describe the experimental setup and results. Table \ref{tab:env_summary} summarises the environments.

\begin{table}[h]
\centering
\caption{Summary of environments. $N$: number of agents. `Bound': proven classical performance limit. `Q. Framework': quantum simulation library.}
\label{tab:env_summary}
\begin{tabular}{llcccl}
\toprule
\textbf{Env.} 
  & \textbf{Type} 
  & $N$ 
  & \textbf{Actions}
  & \textbf{Bound}
  & \textbf{Q.\ Framework} \\
\midrule
CHSH 
  & Cooperative game 
  & 2 
  & Discrete 
  & 0.75 (proven) 
  & Qiskit Aer \\

CoinGame 
  & Mixed benchmark 
  & 2, 4 
  & Discrete 
  & - 
  & TF-Quantum \\
CoopNav 
  & Cooperative benchmark 
  & 2, 3, 4 
  & Discrete 
  & - 
  & TF-Quantum \\
\bottomrule
\end{tabular}
\end{table}

\subsection{CHSH Game}
\label{sec:env_chsh}
The CHSH (Clauser-Horne-Shimony-Holt) game ~\cite{clauser_proposed_1969} is a two-agent cooperative coordination game that serves as a standard testbed \cite{brunner2014bellnonlocality} for quantum advantage. We use it as a calibration benchmark for QMARL because it has a classical limit that is mathematically proven and any quantum agent consistently exceeding it provides unambiguous evidence of quantum advantage.

Two agents, Alice and Bob, receive independent binary inputs $x, y \in \{0, 1\}$ drawn uniformly at random. Each agent produces a binary output, $a \in \{0,1\}$ (Alice) and $b \in \{0,1\}$ (Bob), without communicating. They win the round if and only if $ a \oplus b = x \wedge y$. This requires agents to output the same bit in three out of four input combinations (when $x \cdot y=0$) and \emph{different} bits in one 
combination ($x=y=1$). The full mapping of input pairs to winning action parities is detailed in Appendix ~\ref{appendix:chsh_parity_condition}.

As demonstrated by ~\cite{clauser_proposed_1969}, no classical strategy, \textit{deterministic or probabilistic}, in which Alice acts on $x$ alone and Bob acts on $y$ alone can achieve a win rate exceeding $0.75$ and we posit that any consistently higher win rate is direct evidence of quantum advantage.
\begin{figure}[htbp]
    \centering
    \vspace{-10pt}

    \includegraphics[width=0.95\linewidth]{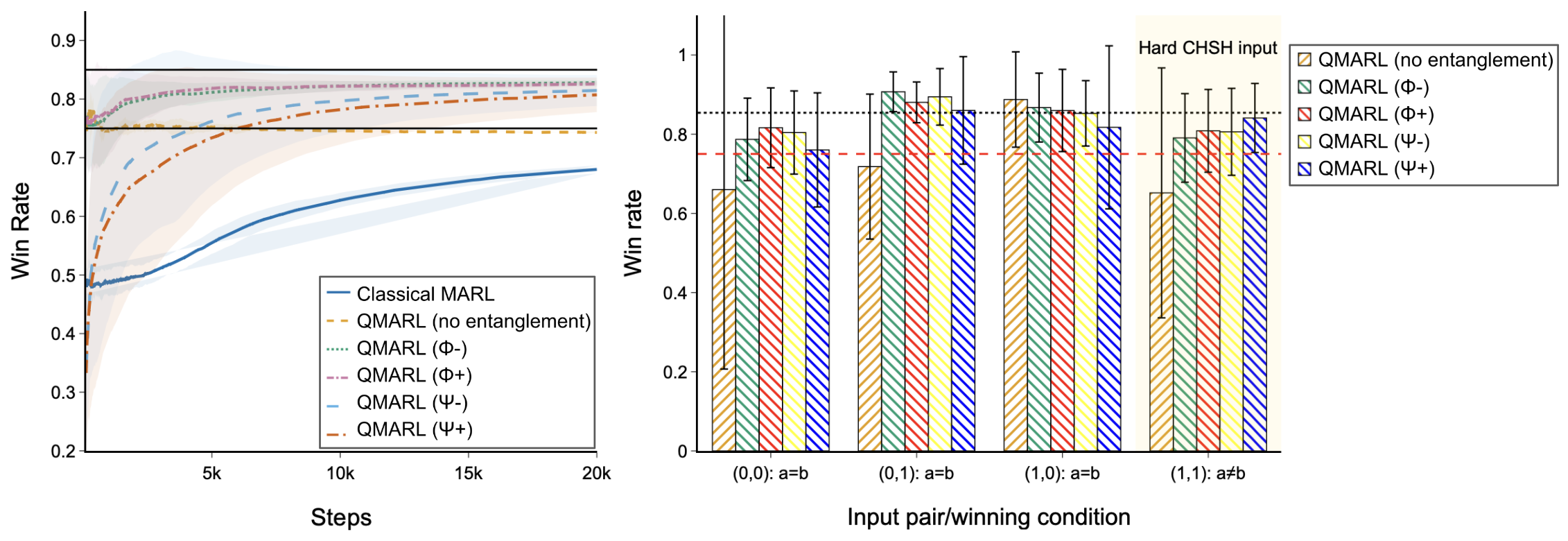}
    \caption{CHSH win rate versus training steps with std dev bands. (a)  Effect of each of the  four Bell states ($|\Phi-\rangle$, $|\Phi+\rangle$, $|\Psi-\rangle$, $|\Psi+\rangle$). (b) Win rate by input pair for all five QMARL variants. The horizontal lines mark the proven classical bound ($0.75$) and the Tsirelson bound ($0.854$).}
    \label{fig:chsh_plot}
\end{figure}

If the agents share a pre-entangled two qubit (one qubit per agent) Bell state (like $|\Phi+\rangle = (|00\rangle + |11\rangle)/\sqrt{2}$), and apply local measurements parameterised by learned rotation angles, the quantum-mechanical correlations can violate the classical bound. Using this quantum approach, the theoretical maximum becomes $\cos^2(\pi/8) \approx 0.854$ win rate 
(Tsirelson bound~\cite{cirelson_quantum_1980}).

We model CHSH as a repeated cooperative multi agent problem. At each step $t$: (i) The environment samples $(x_t, y_t) \sim \mathrm(\{0,1\}^2)$; (ii) Alice observes $x_t$ only; Bob observes $y_t$ only; (iii)  Alice executes policy $\pi_A(a \mid x_t; \theta_A)$ and Bob executes policy $\pi_B(b \mid y_t; \theta_B)$, independently; (iv) Both agents receive a shared scalar reward  $r_t = \mathbf{1}[(a_t \oplus b_t) = (x_t \wedge y_t)]$; and (v) Parameters $\theta_A, \theta_B$ are updated to maximise $\mathbb{E}_t[r_t]$.

We train five agent configurations: classical MARL, QMARL with non entangled agents and QMARL  with each of the four Bell states ($|\Phi+\rangle$, $|\Phi-\rangle$, $|\Psi+\rangle$, $|\Psi-\rangle$) for entangling the two agents. Notably, prior work \cite{gardiner_learning_2026} demonstrated quantum advantage in CHSH using a centralised quantum coordinator where agents act on coordinator outputs rather than purely independently. Our setup is strictly decentralized with each agent using an independent VQC measuring its own share of the pre-prepared entangled state, with no shared model at runtime. This represents true entanglement in a decentralized multi agent system. Detailed experimental setup can be found in Appendix ~\ref{appendix:chsh_experimental_setup}.

Figure ~\ref{fig:chsh_plot} shows win rate versus training steps with standard deviation bands for classical MARL, QMARL without entanglement and QMARL with entanglement variants. Classical MARL converges at and below the classical bound of $0.75$. QMARL without entanglement tracks the classical baseline throughout, confirming that the quantum circuit alone provides no coordination advantage. Only entangled QMARL variants consistently exceeds $0.75$ and approaches the Tsirelson bound of $0.854$, establishing that entanglement is the active mechanism driving quantum advantage.

As we can see, entanglement structure matters: while all entangled variants consistenly exceed the classical bound and outperform/converge faster than classical variant, $|\Psi-\rangle$ and $|\Psi+\rangle$ show greater variance and converge more slowly. Unentangled QMARL (orange dashed) reaches the classical baseline but does not cross it, confirming that the presence of a quantum circuit is insufficient, entanglement structure is critical. 

To understand \textit{where} the quantum advantage originates, in Figure ~\ref{fig:chsh_plot}, we break down win rate by input pair at convergence (with $\beta = 0.0$). All five QMARL variants perform comparably on the three easy pairs: $(0,0)$, $(0,1)$ and $(1,0)$. The critical difference emerges on the $(1,1)$ pair, the only input requiring agents to output \textit{different} actions ($a \neq b$). Unentangled QMARL fails on this pair (win rate $\approx 0.65$, below the classical bound), while all four Bell state variants achieve win rates of $0.75$-$0.85$ on this pair, consistently at or above the classical bound. This confirms that entanglement specifically resolves the coordination constraint that is classically intractable.

Additional results in \ref{appendix:chsh_additional_results} analyse the effect of entropy regularisation and show that all entangled variants have the same effect, regardless of hyperparameter choice (See Figures ~\ref{fig:chsh_entropy_agg})-~\ref{fig:chsh_entropy_bytype}).

\subsection{CoinGame}
CoinGame is a grid world environment with mixed cooperative and competitive rewards, where each agent must collect coins of its own colour while avoiding collecting coins belonging to other agents (which we call \textit{stealing}). We test two variants: CoinGame-2 (CG-2; 2 agents on a 3x3 grid) and CoinGame-4 (CG-4; 4 agents on a 5x5 grid).

\begin{figure}[htbp]
    \centering
    \vspace{-10pt}
    \includegraphics[width=0.95\linewidth]{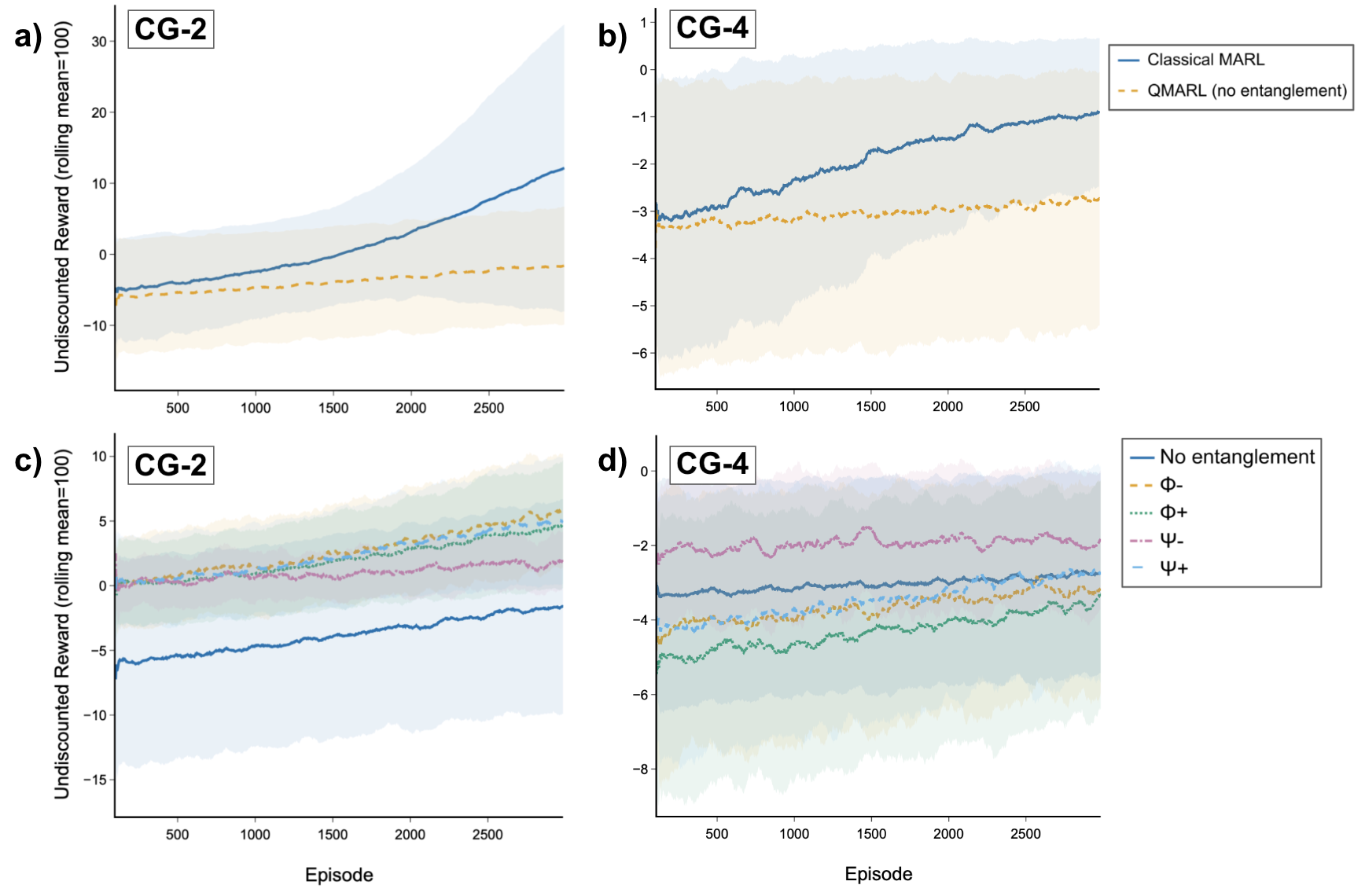}
    \caption{CoinGame: Classical MARL versus QMARL without entanglement (top) and across Bell states (bottom).
    }
    \label{fig:cg2_baseline}
\end{figure}

We first compare classical MAA2C against QMARL with no entanglement to isolate the effect of the quantum policy representation, and then compare all four entanglement variants ($|\Phi-\rangle$, $|\Phi+\rangle$, $|\Psi-\rangle$, $|\Psi+\rangle$) against the no entanglement QMARL baseline (See Figure ~\ref{fig:cg2_baseline}).

Classical MARL achieves higher undiscounted rolling reward than unentangled QMARL in both settings. The gap is most pronounced in CG-2, where the classical policy improves steadily from negative reward to a clearly positive final reward, while unentangled QMARL remains negative throughout training.

The Bell-state comparisons show that entanglement does not have a uniform effect across CoinGame settings. In CG-2, all Bell-state variants outperform the unentangled QMARL baseline, with $|\Phi{-}\rangle$ achieving the best final reward. Whereas in CG-4, $|\Psi{-}\rangle$ has the best trajectory, while all other variants remain below or close to the unentangled baseline. 

Thus, entanglement structure is crucial even in sequential MDP environments. This highlights that the entanglement choice matters and some entanglement states can even actively harm performance. Thus, its effect is environment-dependent rather than uniformly beneficial. 

To further assess the role of circuit size, we performed a qubit-depth ablation over $(n_q,d)\in\{4,6\}\times\{3,4\}$. The ablation results are reported in Appendix ~\ref{appendix:CG_additional_results}, while the full CoinGame environment specification and hyperparameters are provided in Appendix ~\ref{appendix:coingame}.

\subsection{Cooperative Navigation (CoopNav)}
CoopNav is a discrete grid world environment in which $N$ agents must navigate to a shared goal as quickly as possible without colliding. The task is purely cooperative: all agents share the same reward signal. We test $N \in \{2, 3, 4\}$ agents on a $5{\times}5$, $7{\times}7$ and $9{\times}9$  grid. An episode succeeds when any agent reaches the goal, at which point all agents receive $+1$ and the episode ends. The step penalty encourages agents to reach the goal quickly; the collision penalty discourages agents from occupying the same cell. We report \textit{success rate} (fraction of episodes where the goal is reached), collision count and episode length. A stochastic slip probability $p_{\text{slip}} = 0.10$ is applied in all experiments, replacing the intended action with a uniformly random action $10\%$ of the time, to test robustness of learned policies. Unlike CHSH and CoinGame, CoopNav is a purely cooperative task with no competitive incentives. As with CoinGame, there is no known classical performance bound; we present empirical results in this section. 
Detailed environment and experimental setup information is provided in Appendix~\ref{appendix:coopnav_exp_setup}.

\begin{figure}[htbp]
    \centering
    \vspace{-10pt}
    \includegraphics[width=\linewidth]{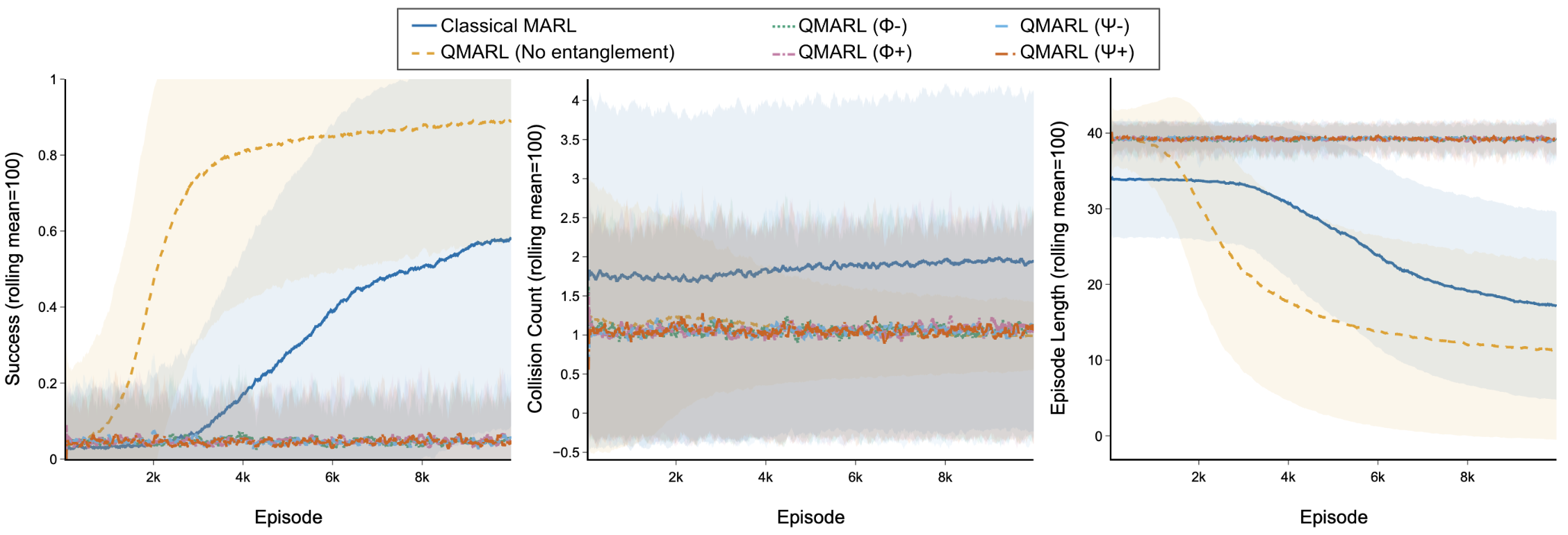}
    \caption{CoopNav: Effect of entanglement variant on success rate, collision count and episode length.
    }
    \label{fig:cn_entanglement}
    \vspace{-10pt}
\end{figure}

We train all variants using MAA2C with CTDE, a centralised critic observing the concatenated observations of all agents, and per-agent actors trained on local observations only. In no entanglement QMARL variant, all quantum actor variants have \textit{no inter agent entanglement}; each agent's VQC operates independently on its local observation. The advantage therefore arises from the quantum policy representation itself, not from entanglement-based coordination.

Figure ~\ref{fig:cn_entanglement} compares Classical MAA2C with unentangled QMARL and entangled QMARL variants on CoopNav. The unentangled QMARL variant achieves the strongest performance across the main task metrics, whereas Classical MAA2C improves more slowly and but plateaus at a lower final success rate. The unentangled QMARL policy also reduces episode length substantially compared to the classical baseline. Collision counts are lower for all quantum variants than for Classical MAA2C, although this metric must be interpreted together with success rate because the entangled variants largely fail to solve the navigation task.

This finding contrasts directly with CHSH, where entanglement was essential. In CoopNav, entanglement appears to add noise rather than coordination signal in this setting, interfering with policy learning. The quantum advantage in CoopNav is therefore driven entirely by the expressiveness of the VQC policy representation.

\begin{figure}[htbp]
    \centering
    \vspace{-10pt}
    \includegraphics[width=\linewidth]{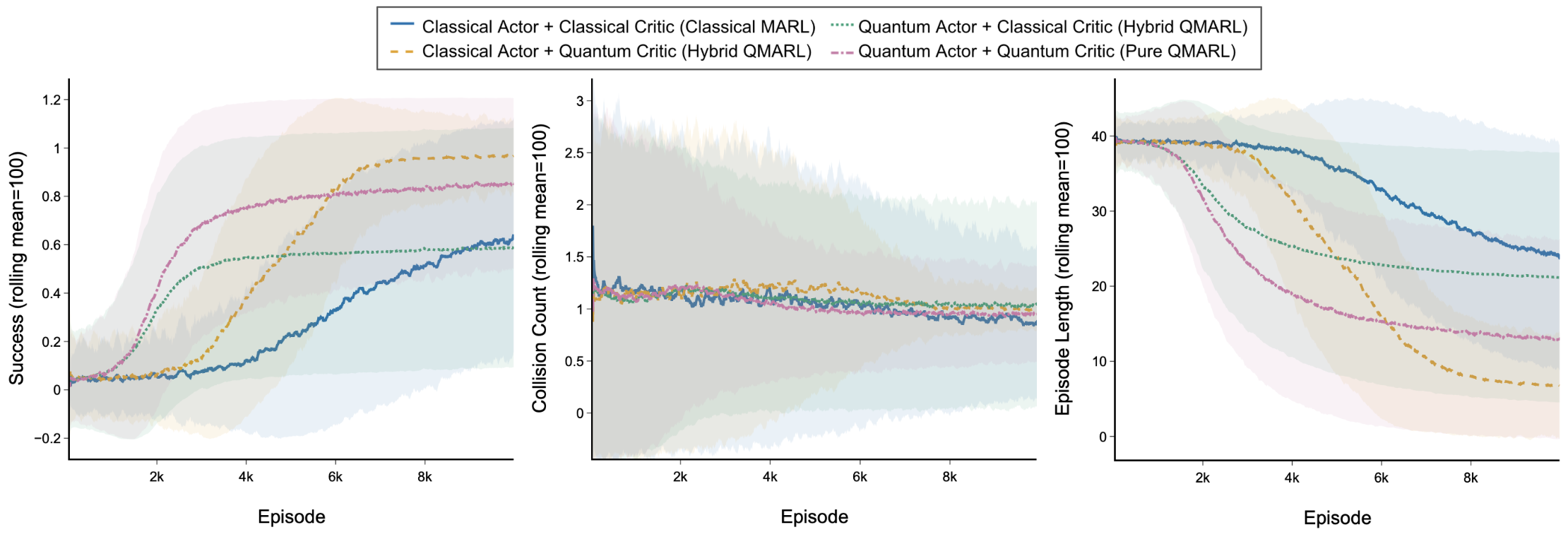}
    \caption{CoopNav: Effect of  quantum hybridizations on  success rate, collision count and episode length. 
    }
    \label{fig:cn_baseline}
\end{figure}

Next, Figure ~\ref{fig:cn_baseline} compares the four actor-critic configurations. Classical Actor$+$Quantum Critic reaches the strongest final performance, but it learns slowly. Pure QMARL converges earlier and remains competitive, while Quantum Actor$+$Classical Critic is the most stable across configurations. Classical MARL is consistently the weakest baseline. Appendix Tables~\ref{tab:cn_ablation}-\ref{appendix:cn_actor_critic_summary} confirm that the most reliable overall design is Hybrid QMARL with a Quantum Actor and Classical Critic; high aggregate performance with low variance. Pure QMARL can perform well but is less stable, whereas Classical Actor$+$Quantum Critic benefits from a strong late training trajectory but lags behind across the full training run. Overall, CoopNav suggests that placing the quantum component in the decentralised actor is more reliable than using it only in the centralised critic.

\section{Discussion and Conclusion}
Our experiments across three environments reveal two mechanistically distinct forms of quantum benefit in MARL and one important negative finding that limits the scope of our claims. Several limitations bound the 
results discussed below. All experiments are conducted on noiseless quantum simulators. Real quantum hardware introduces decoherence, gate errors, readout noise and finite-shot estimation that are not modelled here. The entanglement advantage demonstrated in CHSH depends on high-fidelity quantum state preparation; under realistic noise, this fidelity degrades and the advantage may shrink or vanish. Noise-aware evaluation and real quantum hardware experiments are necessary before any claim of practical quantum advantage.

CHSH is not a sequential MDP as it has no evolving environment state, no temporal credit assignment, and no sequential decision making. It is a one-shot game where each step is independent. The entanglement advantage demonstrated here is therefore specific to nonlocal game structures with provable classical bounds. Whether equivalent advantages can be achieved in genuine MDPs, where temporal dynamics, credit assignment and partial observability interact, remains an open question.

In CoinGame and CoopNav, it is difficult to isolate quantum specific effects from differences in policy architecture or training dynamics.  Also, all experiments use small agent counts ($N \leq 4$) and small grids. Whether the quantum advantages observed here persist, grow or vanish as the number of agents and environment complexity scale is unknown. Simulating large numbers of qubits on classical hardware is expensive,  severely limiting the scale at which QMARL can be evaluated in simulation.

In CHSH, entanglement is the mechanism that produces quantum advantage. Unentangled QMARL tracks the classical baseline, while all four entangled variants consistently exceed the classical bound. The per-input-pair analysis makes the mechanism explicit: the advantage is entirely concentrated on the $(1,1)$ pair, the only input requiring agents to output different actions. 
No classical strategy, regardless of capacity or training budget, can resolve this constraint; entanglement does so by establishing non-local correlations that have no classical justification/parallel. 
Moreover, entanglement structure matters: some entanglement states yield consistent gains, while others show higher variance and can harm performance, so an adaptive approach needs to be employed when using them.

CoopNav shows a different form of quantum benefit. Inter-agent entanglement does not improve performance; instead, the no-entanglement quantum actor variant consistently outperforms the entangled variants. Here, the empirical advantage arises from the VQC as a policy function approximator rather than from non-local quantum correlations between agents. 
The quantum circuit acts as a compact policy representation whose structure appears well suited to cooperative navigation. The actor-critic ablation further shows that quantum placement matters: the most reliable overall design is the hybrid QMARL configuration with a quantum actor and classical centralised critic, while pure QMARL can perform well but is less stable, and a quantum critic alone learns more slowly despite strong late-stage performance.

CoinGame provides a more cautionary finding as Classical MAA2C outperforms unentangled QMARL on undiscounted reward. The quantum circuit did not successfully learn the task structure in CoinGame within the training budget. Also, no entangled variant matches classical performance.

Across all environments, quantum variants use comparable or fewer trainable parameters than classical baselines. In CHSH the comparison is exact (4 versus 4 parameters); in CoinGame and CoopNav quantum actors use comparable or fewer parameter counts with the majority parameters residing in the classical preprocessing layer. Model capacity therefore cannot explain the observed performance differences, either the advantages in CHSH and CoopNav.

These results together establish a clear picture; quantum advantage in MARL can arise from two fundamentally different sources: entanglement-based coordination in problems with classically intractable joint constraints and VQC policy expressiveness. These are different mechanisms. The first requires a specific problem structure, a provable classical ceiling that entanglement can breach. The second is a property of the quantum circuit as a function class, independent of explicit inter-agent quantum correlations. Identifying which mechanism is relevant to a given problem is therefore a necessary step in any QMARL application. Notably, the parameter analysis reveals that the majority of trainable parameters in our quantum variants/actors reside in the classical preprocessing and readout layers rather than in the VQC itself (Appendix Table~\ref{tab:param_counts}). The VQC core is extremely compact, yet it drives the performance gains. This suggests that data encoding and classical pre- and post-processing are significant bottlenecks in current hybrid quantum-classical architectures. More expressive or task-specific encoding schemes that better exploit the geometry of state space may unlock/enable stronger quantum advantage. Improving the interface between classical observations and quantum circuits is therefore a key direction for future QMARL research.

\bibliographystyle{unsrt}  
\bibliography{references}

\medskip


\appendix

\section{Appendix A: CHSH}
\label{appendix:chsh}

\subsection{CHSH Parity Condition}
\label{appendix:chsh_parity_condition}
The formal payoff structure and required action parities for all input combinations in the CHSH game are detailed in Table ~\ref{tab:chsh_parity_table}.

\begin{table}[h!]
\centering
\caption{CHSH Game Payoff and Parity Table.}
\label{tab:chsh_parity_table}
\begin{tabular}{p{1.48cm} p{1.48cm} p{1.48cm} p{2.48cm} p{2.48cm} r}
\toprule
\textbf{Agent 1 ($x$)} 
  & \textbf{Agent 2 ($y$)} 
  & \textbf{Target ($x \wedge y$)} 
  & \textbf{Required Parity ($a \oplus b$)} 
  & \textbf{Winning Actions ($a, b$)} 
  & \textbf{Reward} \\
\midrule
0 & 0 & 0 & Match (0)  & $(0, 0)$ or $(1, 1)$ & 1 \\
0 & 1 & 0 & Match (0)  & $(0, 0)$ or $(1, 1)$ & 1 \\
1 & 0 & 0 & Match (0)  & $(0, 0)$ or $(1, 1)$ & 1 \\
1 & 1 & 1 & Differ (1) & $(1, 0)$ or $(0, 1)$ & 1 \\
\bottomrule
\end{tabular}
\end{table}

\subsection{CHSH: Experimental Setup}
\label{appendix:chsh_experimental_setup}
For experimental details in the Section~\ref{sec:env_chsh}, hyperparameters are summarised in Table ~\ref{tab: chsh_hyperparams}. All QMARL variants use 1 qubit per agent and circuit depth 1. We use REINFORCE method with a moving average baseline, and train for 20000 steps and evaluate win rate over 1000 episodes.  

All experiments run with entropy coefficients $\beta \in \{0.0, 0.2\}$ to assess sensitivity to entropy regularisation. Entropy regularisation adds an extra term ($H(\pi_i) = -\sum_a \pi_i(a)\log\pi_i(a)$) to the REINFORCE update function that enables/encourages more exploration during training by rewarding the policy for uncertainty, also avoiding getting stuck in local minima. With $\beta = 0$, there is no entropy regularisation. Primary results have fixed $\beta$ as 0.

\begin{table}[h]
\centering
\caption{CHSH experimental hyperparameters.}
\label{tab: chsh_hyperparams}
\begin{tabular}{ll}
\toprule
\textbf{Hyperparameter} & \textbf{Value} \\
\midrule
Algorithm           & REINFORCE \\
Learning rate       & 0.02 \\
Baseline momentum   & 0.95 \\
Training steps      & 20000 \\
Evaluation episodes & 1000 \\
Qubits per agent    & 1 \\
Circuit depth       & 1 \\
Entropy coefficient $\beta$ & \{0.0, 0.2\} \\
Seeds               & 10 \\
Simulator           & Qiskit Aer 0.17.2 (statevector, noiseless) \\
\bottomrule
\end{tabular}
\end{table}

\subsection{CHSH: Additional Results}
\label{appendix:chsh_additional_results}
\begin{figure}[htbp]
    \centering
    \includegraphics[width=0.84\linewidth]{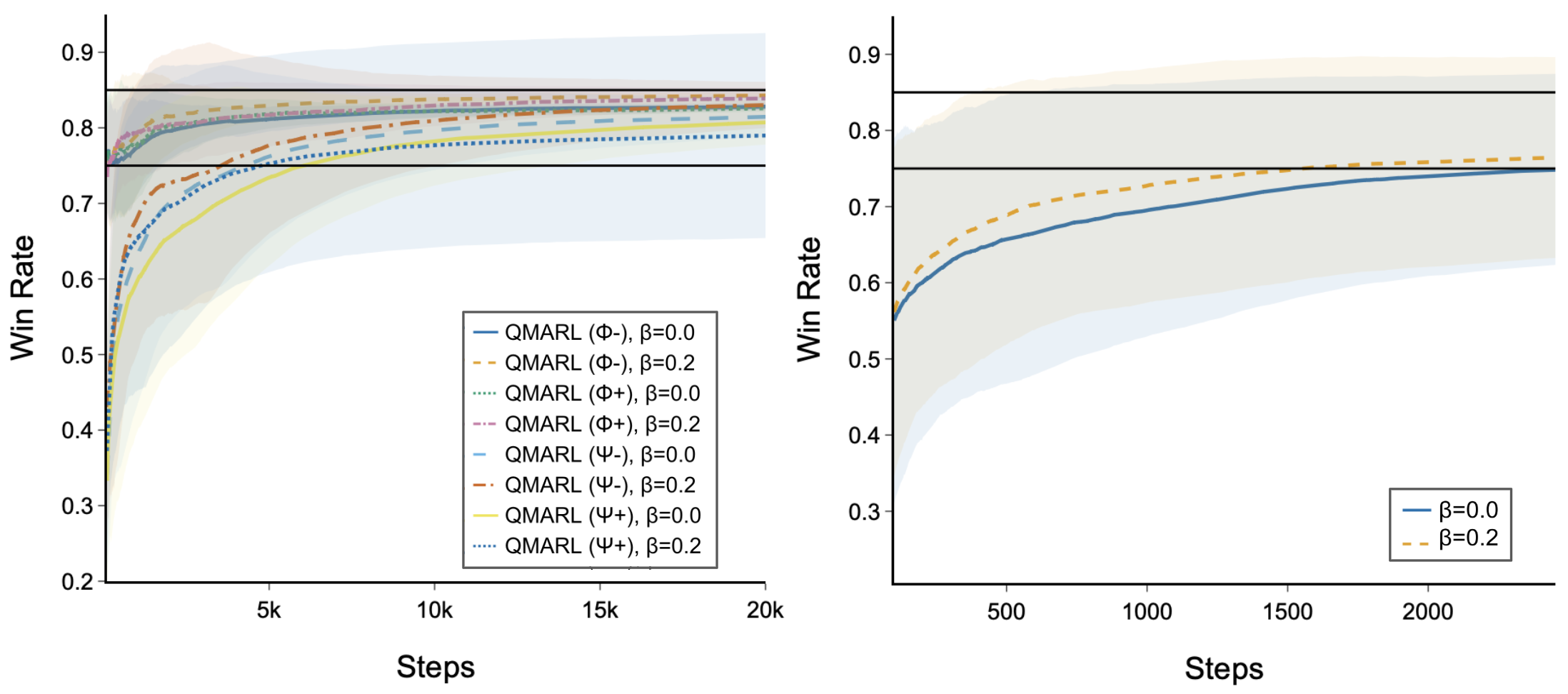}
    \caption{
        Effect of entropy regularisation. 
        Left panel shows win rate curves for all eight combinations of entanglement state and entropy coefficient ($\beta \in \{0.0, 0.2\}$). 
        Right panel aggregates win rate averaged over all four entangled variants at $\beta = 0.0$ (blue) versus $\beta = 0.2$ (orange dashed).}
    \label{fig:chsh_entropy_agg}
\end{figure}

Here, we  examine the effect of entropy regularisation.

\begin{figure}[htbp]
    \centering
    \includegraphics[width=0.84\linewidth]{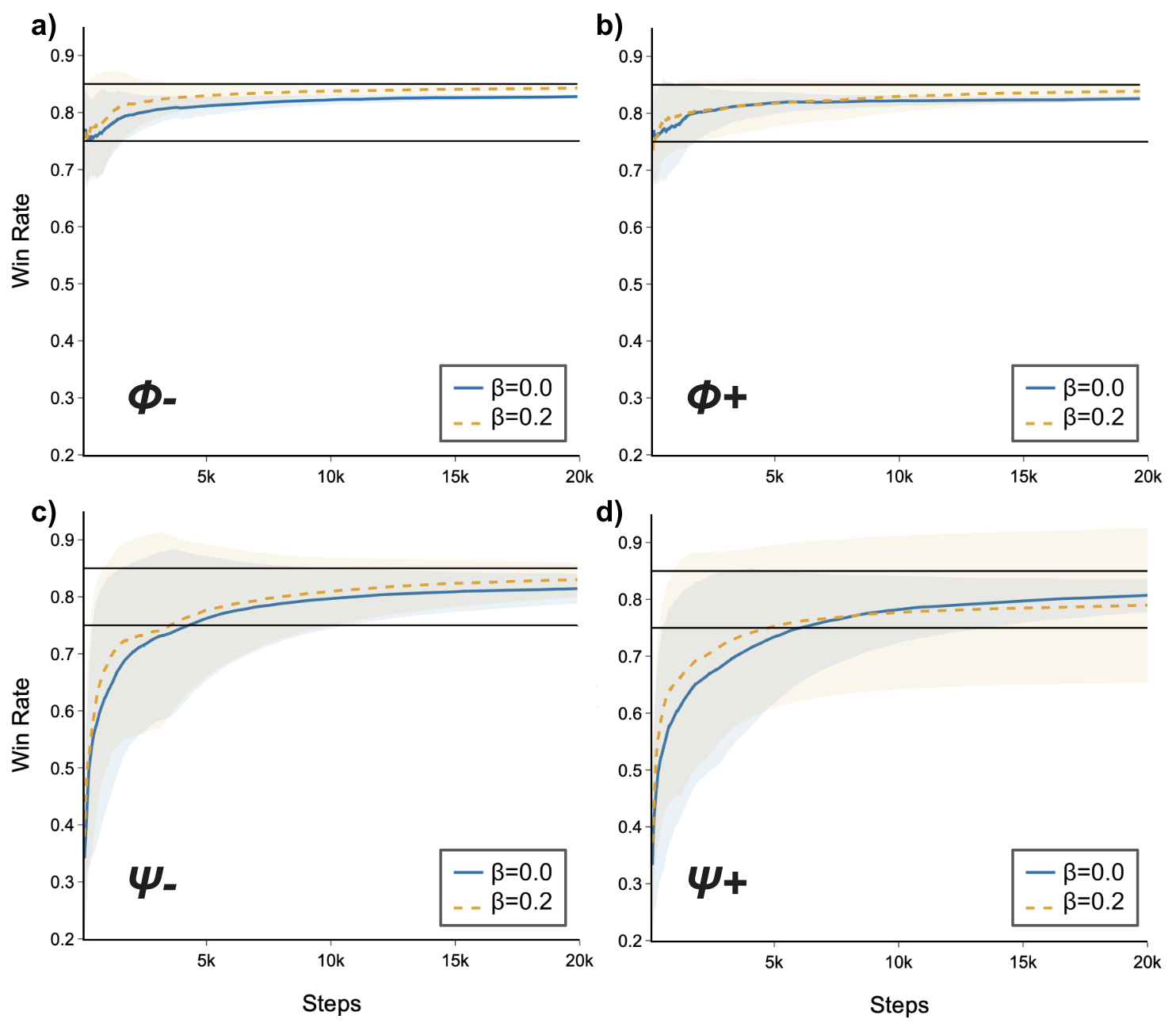}
    \caption{
        Effect of entropy regularisation per entanglement type. Win rate versus training steps for $\beta = 0.0$ (blue) versus $\beta = 0.2$ (orange dashed) for each Bell state separately: a)~$|\Phi-\rangle$, b)~$|\Phi+\rangle$, c)~$|\Psi-\rangle$, d)~$|\Psi+\rangle$. 
        The solid black horizontal lines mark the classical bound ($0.75$) and the Tsirelson bound ($0.854$). Shaded bands show std dev across seeds.
        }
    \label{fig:chsh_entropy_bytype}
\end{figure}

Figure ~\ref{fig:chsh_entropy_agg} shows the performance across all eight variant$\times\beta$ combinations, alongside the aggregated results for all entangled variants at $\beta = 0.0$ versus $\beta = 0.2$. The overall effect is small; while $\beta = 0.2$ induces marginally faster convergence in early training, both conditions achieve similar final win rates. Figure ~\ref{fig:chsh_entropy_bytype} confirms this holds true across all four Bell states individually; none show a meaningful difference between the two entropy coefficients at convergence. Results in Figure ~\ref{fig:chsh_plot} 
are therefore not sensitive to this hyperparameter choice.

\section{Appendix B: CoinGame}
\label{appendix:coingame}

\subsection{Coingame: Experimental Setup}
\label{appendix:coingame_experimental_setup}

In CoinGame, each agent receives a partial observation, a binary $4 \times W \times H$ tensor encoding its own position, other agents' positions, its own coin number information and other coins, as a vector of length $4WH$ as the input. Each agent has four discrete moves as actions (N/S/E/W); the agent stays in place at boundaries. Rewards are individual: $+1$ for collecting any coin, and $-2$ to the coin owner if their coin is stolen. Each episode runs for 150 steps with one coin active at a time, respawning after each collection.

In CoinGame experiments, we train all variants using MAA2C with CTDE, using a centralised classical critic that observes the concatenated observations of all agents, together with separate per-agent actors. The classical actor is a fully connected feedforward neural network with layer sizes $36 \rightarrow 12 \rightarrow 4$ (Dense $36 \to 12 \to 4$, ReLU activations, softmax output) and the centralised critic is a fully connected neural network with layer sizes $72 \rightarrow 12 \rightarrow 1$ (Dense $72 \to 12 \to 1$, ReLU activations, linear output). For the quantum variant, the classical actor network is replaced by a variational quantum circuit (VQC) with $n_q$ qubits and depth $d$. The measurement outcomes are then passed through a classical fully connected readout layer of size $2^{n_q} \rightarrow 4$ (Dense $2^{n_q} \to 4$) to produce action logits. The centralised critic is kept identical across classical and quantum variants.

This setting is referred to as \textit{hybrid QMARL}, as defined in Section~\ref{sec:methods}, because the actor is quantum while the critic and optimisation pipeline remain classical. 
Parameter accounting for the classical and quantum variants is reported separately in Table ~\ref{tab:param_counts}, including preprocessing, VQC, readout, actor, critic and total trainable parameters.

\begin{table}[htbp]
\centering
\caption{Hyperparameters for CoinGame experiments.}
\label{tab:cg_hyperparams}
\begin{tabular}{lll}
\toprule
\textbf{Hyperparameter} & \textbf{Value} & \textbf{Note} \\
\midrule
Actor learning rate      & $3 \times 10^{-4}$ & Adam \\
Critic learning rate     & $1 \times 10^{-3}$ & Adam \\
Discount factor $\gamma$ & $0.95$             & \\
Episode length           & $150$ steps        & \\
Training episodes (CG-2) & $3000$          & \\
Training episodes (CG-4) & $5000$          & \\
Seeds                    & $10$               & \\
Action slip prob.\       & $0.0$              & Deterministic \\
Steal penalty            & $-2.0$             & \\
\bottomrule
\end{tabular}
\end{table}

\subsection{CoinGame: Additional Results} 
\label{appendix:CG_additional_results}

\begin{figure}[htbp]
    \centering
    \includegraphics[width=0.84\linewidth]{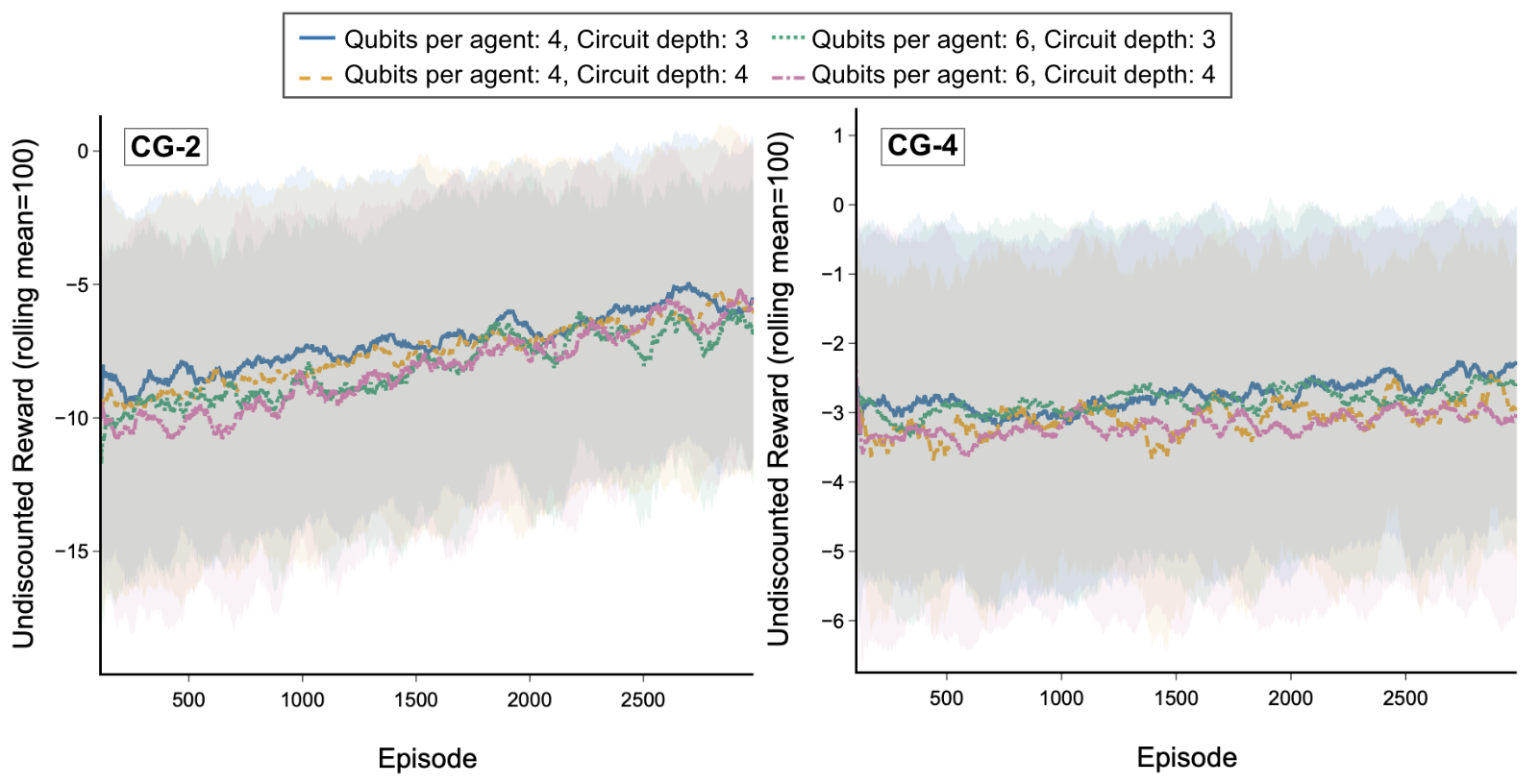}
    \caption{CoinGame: Qubit count and circuit depth ablation (no entanglement). 
    Shaded bands show std dev across 10 seeds.
    }
    \label{fig:cg2_qubit_depth}
\end{figure}

Figure ~\ref{fig:cg2_qubit_depth} shows the qubit/depth ablation for QMARL without entanglement. 

All four configurations (Q4-D3, Q4-D4, Q6-D3, Q6-D4) perform similarly, clustering tightly with high variance throughout training. No configuration clearly dominates on either metric. The rolling means show marginal differences within noise bounds.  
Figure ~\ref{fig:cg2_qubit_depth} shows the qubit/depth ablation for CG-4. As in CG-2, all configurations cluster tightly with high variance and no dominant configuration emerges on either metric.

\section{Appendix C: CoopNav}
\label{appendix:coopnav}
\subsection{CoopNav: Experimental Setup}
\label{appendix:coopnav_exp_setup}

In CoopNav, $N$ agents navigate a shared grid world toward a common goal while avoiding collisions. We test $N \in \{2,3,4\}$ agents on $5\times5$, $7\times7$, and $9\times9$ grids. Each agent observes a normalised continuous vector of shape $2(N{+}1)$ containing its own position, the positions of all other agents, and the goal position, with all coordinates divided by the grid size to lie in $[0,1]$. Each agent has five discrete actions: N/S/E/W/Stay. The shared reward at each step is:
\begin{equation}
    r^t = \underbrace{-0.01}_{\text{step penalty}}
        + \underbrace{-0.05 \cdot
          {1}[\text{collision}]}_{\text{collision penalty}}
        + \underbrace{+1.0 \cdot
          {1}[\text{goal reached}]}_{\text{success reward}}
    \label{eq:cn_reward}
\end{equation}
An episode terminates successfully when any agent reaches the goal, at which point all agents receive the shared success reward.

The classical actor is a feedforward neural network (Dense $2(N{+}1) \to 16 \to 16 \to 5$ layers, ReLU activations, softmax output); the centralised critic is a feedforward neural network (Dense $2N(N{+}1) \to 32 \to 16 \to 1$, ReLU activations, linear output). For quantum variants, the classical actor is replaced by a VQC with $n_q$ qubits and circuit depth $d$; measurement outcomes are passed through a classical readout layer (Dense $2^{n_q} \to 5$) to produce action logits. Parameter accounting for all classical and quantum variants is reported in Table ~\ref{tab:param_counts}. All experimental results are averaged over 10 random seeds, and smoothed curves use a rolling-mean window of 100. Hyperparameters are reported in Table ~\ref{tab:cn_hyperparams}.

\begin{table}[htbp]
\centering
\caption{Hyperparameters for CoopNav experiments.}
\label{tab:cn_hyperparams}
\begin{tabular}{lll}
\toprule
\textbf{Hyperparameter} & \textbf{Value} & \textbf{Note} \\
\midrule
Actor learning rate      & $2 \times 10^{-4}$ & Adam \\
Critic learning rate     & $3 \times 10^{-4}$ & Adam \\
Discount factor $\gamma$ & $0.99$             & \\
Max episode steps        & $40$               & \\
Training episodes        & $10000$         & \\
Seeds                    & $10$               & \\
Slip probability $p_{\text{slip}}$ & $0.10$  & Random action \\
$N$ agents               & $2, 3, 4$          & \\
Grid size                & $5{\times}5$, $7{\times}7$, $9{\times}9$       & \\
\bottomrule
\end{tabular}
\end{table}

\begin{figure}[htbp]
    \centering
    \includegraphics[width=\linewidth]{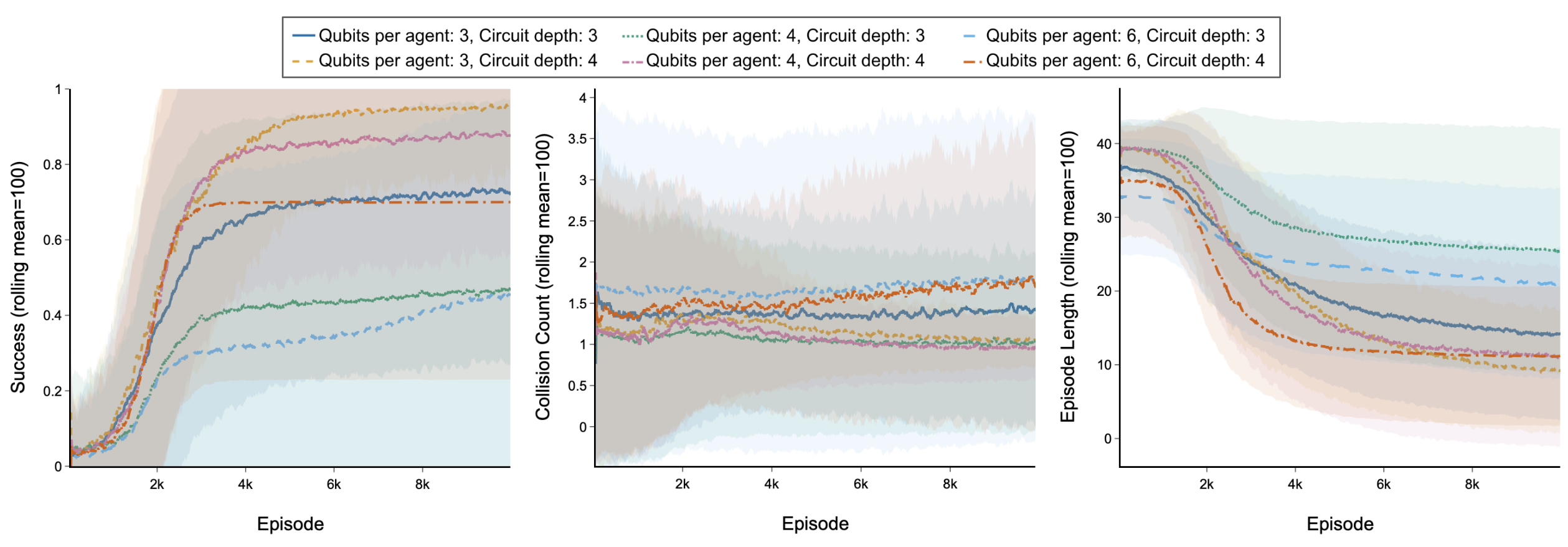}
    \caption{CoopNav: Qubit count and circuit depth ablation. 
    Shaded bands show std dev across 10 seeds.
    }
    \label{fig:cn_qubit_depth}
\end{figure}

\subsection{CoopNav: Additional Results}
\label{appendix:coopnav_additional_results}
Figure~\ref{fig:cn_qubit_depth} shows the qubit-depth ablation for six unentangled quantum actor configurations. Unlike CoinGame, where the qubit-depth variants cluster closely, CoopNav shows clearer sensitivity to circuit size. The larger six-qubit configurations reach higher success-rate plateaus and shorter episode lengths, while the smallest configurations are weaker. Even collision counts show less separation than success rate and episode length, indicating that circuit size primarily affects navigation success and path efficiency rather than collision avoidance alone.

Tables ~\ref{tab:cn_ablation}-~\ref{appendix:cn_actor_critic_summary} provide the full actor-critic ablation results, reporting performance across qubit and depth settings and aggregated by actor-critic type. Together with Figure ~\ref{fig:cn_baseline}, these results confirm that the most reliable overall design is the hybrid QMARL setting with a quantum actor and classical centralised critic. Pure QMARL can also perform well, but its performance varies more across configurations. The classical-actor/quantum-critic variant reaches strong late-stage performance in the learning curves, but is less reliable when averaged across the full training run.

The CoopNav ablations therefore support a different conclusion from CHSH. In CHSH, entanglement is the mechanism that enables formal quantum advantage. In CoopNav, inter-agent entanglement does not improve performance over the unentangled quantum actor baseline. The empirical advantage in CoopNav is instead associated with the VQC actor as a policy representation and with the stability provided by retaining a classical centralised critic.

\input{Tables/CoopNav_Ablation_v1.0}

\begin{table}[h]
\centering
\caption{CoopNav actor-critic variant ablation: mean success rate averaged over all training episodes and seeds ($5{\times}5$ grid, $p_\text{slip}=0.10$, 10 seeds).}
\label{appendix:cn_actor_critic_summary}
\begin{tabular}{llllrrrr}
\toprule
\textbf{Hybridisation} & \textbf{Actor} & \textbf{Critic} 
  & \textbf{SR (Mean)} & \textbf{Std} & \textbf{SR (Best)} \\
\midrule
Hybrid QMARL  & Quantum   & Classical & \textbf{0.817} & 0.036 & 0.903 \\
Pure QMARL    & Quantum   & Quantum   & 0.654          & 0.254 & 0.871 \\
Hybrid QMARL  & Classical & Quantum   & 0.549          & 0.140 & 0.726 \\
Classical MARL & Classical & Classical & 0.338          & 0.198 & 0.615 \\
\bottomrule
\end{tabular}
\end{table}

\section{Appendix D: Learning Framework}
\label{appendix:learning_framework}

\paragraph{Centralised Training, Decentralized Execution (CTDE):} Training decentralized agents is difficult as each agent only sees part of the environment, yet its actions affect everyone. CTDE ~\cite{lowe_multi-agent_2020} resolves this by splitting training and execution into two phases:

\begin{itemize}
    \item Training (centralised): a single critic network $V_\Phi(s)$ has access to the full global state $s^t$ concatenating all agents' local observations, and estimates the expected total future reward for the agents team. This global view provides a richer training signal than any single agent's local observation alone.

    \item Execution (decentralized): each agent's actor $\pi_i$ uses only its own local observation $o_i^t$. The critic is used only during training and is discarded at execution time so that the agents act independently with no communication.
\end{itemize}

\paragraph{Multi Agent Advantage Actor Critic (MAA2C):} We use MAA2C ~\cite{mnih_asynchronous_2016, eQMARL} as the training algorithm. It has two components that are updated together after each episode:

1. The critic learns to estimate how much total future reward the team can expect from the current state, and is trained to minimise the prediction error (TD error) between its estimate and what actually happened:
\begin{equation}
    \mathcal{L}_{\text{critic}}(\Phi)
    = \underbrace{\left(
        \overbrace{r^t + \gamma V_\Phi(s^{t+1})}^{
        \text{actual}}
        -\,
        \overbrace{V_\Phi(s^t)}^{
        \text{predicted}}
      \right)^2}_{\text{prediction error to minimise}}
    \label{eq:critic_loss}
\end{equation}

2. Each actor learns a better policy by taking actions that led to better than expected outcomes more often. The advantage function $\hat{A}^t = r^t + \gamma V_\Phi(s^{t+1}) - V_\Phi(s^t)$ measures whether the outcome was better or worse than the critic predicted. Then, the policy is updated to increase the probability of good actions as:
\begin{equation}
    \nabla_{\theta_i} \mathcal{L}(\theta_i)
    = -\mathbb{E}\!\left[
        \underbrace{\hat{A}^t}_{\text{was it good?}}
        \cdot\,
        \nabla_{\theta_i}
        \log \underbrace{
          \pi_i(a_i^t \mid o_i^t; \theta_i)
        }_{\text{probability of action taken}}
    \right]
    \label{eq:actor_loss}
\end{equation}

In our setup, we experiment with combinations of classical and quantum actors and critics: fully classical (classical actor + classical critic), hybrid QMARL (quantum actor + classical critic or classical actor + quantum critic) and pure QMARL (quantum actor + quantum critic).

\paragraph{Special case:} The CHSH game (see Section ~\ref{sec:env_chsh}) is a single-step problem with no temporal dynamics, thus there is no next state to predict. A learned critic is therefore unnecessary.

We use REINFORCE ~\cite{williams_simple_1992} with a simple moving-average baseline $b_t$ instead:
\begin{equation}
    b_t = m \cdot b_{t-1} + (1-m) \cdot r^t,
    \qquad
    \theta_i \leftarrow \theta_i
    + \alpha (r^t - b_t)
    \nabla_{\theta_i}
    \log \pi_i(a_i^t \mid o_i^t; \theta_i)
    \label{eq:reinforce}
\end{equation}
where $\alpha$ is the learning rate and $m$ is the baseline momentum. The baseline $b_t$ estimates the average reward, so $(r^t - b_t)$ measures whether this step was above or below average.

\section{Appendix E: Parameter count methodology and interpretation}
\label{app:parameter_counts}

\begin{table}[H]
\centering
\caption{Trainable parameter counts for all policy architectures. Classical: fully connected Dense layers, params $= n_\text{in} \times n_\text{out} + n_\text{out}$ per layer.
Quantum actor: three components, (i) preprocessing Dense$(n_\text{obs} \to n_q{\times}3)$ [dominant cost], (ii) PQC variational+encoding layers [compact],(iii) readout. Bell state preparation is a fixed non-parametric resource with zero trainable parameters. The centralised critic is identical across classical and quantum 
variants.}
\label{tab:param_counts}
\resizebox{\linewidth}{!}{%
\begin{tabular}{lllrrrrr}
\toprule
\textbf{Env.}
  & \textbf{Variant}
  & \textbf{Preproc.}
  & \textbf{VQC$^\dagger$}
  & \textbf{Readout}
  & \textbf{Actor}
  & \textbf{Critic}
  & \textbf{Total} \\
\midrule
\multirow{2}{*}{CHSH}
  & Classical (Bernoulli)
    & - & - & - & 4 & - & 4 \\
  & QMARL (any Bell, 1q)
    & - & 4 & - & 4 & - & 4 \\
\midrule
\multirow{2}{*}{CG-2}
  & Classical MAA2C
    & - & - & - & 496 & 889 & 1385 \\
  & Hybrid QMARL (4q)
    & 444 & ${\sim}32$--$44$ & 4 & ${\sim}480$--$492$ & 889 & ${\sim}1{,}369$--$1{,}381$ \\
\multirow{2}{*}{CG-4}
  & Classical MAA2C
    & - & - & - & 1264 & 2413 & 3677 \\
  & Hybrid QMARL (4q)
    & 1212 & ${\sim}32\text{--}44$ & 4 
    & ${\sim}1248\text{--}1260$ & 2413 
    & ${\sim}3661\text{--}3673$ \\
\midrule
\multirow{4}{*}{CoopNav}
  & Classical MAA2C
    & - & - & - & 1573 & 5377 & 6950 \\
  & Hybrid QMARL (3q)
    & 684 & ${\sim}36$-$45$ & 15 & ${\sim}735$--$744$ & 5377 & ${\sim}6112$-$6121$ \\
  & Hybrid QMARL (4q)
    & 912 & ${\sim}48$-$60$ & 20 & ${\sim}980$--$992$ & 5377 & ${\sim}6357$--$6369$ \\
  & Hybrid QMARL (6q)
    & 1368 & ${\sim}72$--$90$ & 30 & ${\sim}1470$--$1488$ & 5377 & ${\sim}6847$--$6865$ \\
\bottomrule
\end{tabular}}

\vspace{2pt}

\footnotesize
$^\dagger$VQC count range: $n_q{\times}d{\times}4$ (lower bound) to $(n_q{\times}d{\times}4 + n_q{\times}3)$ (upper bound), depending on whether one extra variational block is prepended before encoding.
\end{table}

\paragraph{Variational Quantum Circuits (VQC):} A VQC (actor) consists of three components, each contributing differently to the parameter count:

1. State encoding/Data encoding (non-parametric): Before the quantum circuit can process the agent's observation from the classical environment $o_i \in \mathbb{R}^{n_\text{obs}}$, the classical data must be mapped into a quantum state. We use \textit{angle encoding}; each component of $o_i$ is used directly as the rotation angle of a single-qubit gate applied to a qubit initialised in state $|0\rangle$. Here, the encoding gates are not learned; their angles are set directly from the observation values at each forward pass, analogous to normalisation ~\cite{weigold_expanding_2021}.

2. Parameterised quantum circuit (PQC): The PQC is the learned core of the quantum network (actor). Conceptually, it plays the same role as the hidden layers of a classical neural network; it transforms the encoded input into a representation from which an action distribution can be computed. A PQC of depth $d$ and $n_q$ qubits consists of $d$ repeated \textit{variational layers}. Each layer has two parts:

\begin{itemize}
    \item Single-qubit rotation gates (parameterised): applied independently to each qubit. These are the \textit{trainable} part of the circuit. Each gate rotates the qubit's quantum state by a learnable angle.
    
    \item Entangling gates (non-parametric): applied between qubits to create quantum correlations. We use CNOT gates in fixed nearest neighbours. These gates have no trainable parameters but are what allow the circuit to capture dependencies 
    between qubits.
\end{itemize}

A qubit's state can be visualised as a point on the surface of a unit sphere (Bloch sphere). Rotation gates move this point along the sphere. The three standard rotation gates are:
\begin{itemize}
    \item $R_Z(\theta)$: rotates the qubit state around the Z-axis of the Bloch sphere by angle $\theta$
    \item $R_Y(\theta)$: rotates around the Y-axis by angle $\theta$
    \item $R_X(\theta)$: rotates around the X-axis by angle $\theta$
\end{itemize}

Any arbitrary single-qubit rotation can be broken down into a a composition of three such rotations. We use the $R_Z$-$R_Y$-$R_Z$ decomposition, meaning three trainable angles are applied per qubit \textit{before} the entangling gates and another three \textit{after}. Thus, the total number of parameters in a VQC becomes:
\begin{equation}
    \text{VQC params} = n_q \times d \times 6
\end{equation}
where the factor of 6 arises from $2 \times 3$ rotations per qubit per layer (two sets of $R_Z$-$R_Y$-$R_Z$, one on each side of the 
CNOT block).

3. Classical readout layer: Once the PQC has processed the input, we need to extract a classical action probability vector from the quantum state. We measure a quantum system using \textit{computational basis}, that is, checking if it is in state $|0\rangle$ or $|1\rangle$? For $n_q$ qubits, there are $2^{n_q}$ possible measurement outcomes (bitstrings). The probability of each outcome is given by the squared magnitude of the corresponding amplitude in the quantum state. This gives a probability distribution over $2^{n_q}$ outcomes (Born's rule). Crucially, this is a deterministic function of the circuit parameters and the input (in simulation); in our experiments, we compute the exact probability vector rather than sampling.

The $2^{n_q}$-dimensional probability vector is then passed through a trainable classical neural network layer (no bias), with total parameters:
\begin{equation}
    \text{Readout params} = 2^{n_q} \times |\mathcal{A}|
\end{equation}
where $|\mathcal{A}|$ is the number of actions. This layer maps quantum measurement probabilities to action logits, from which a softmax gives the action distribution. 

Interestingly, for larger qubit counts, the readout layer can contain more parameters than the VQC itself. This is a known feature of hybrid quantum-classical architectures and is fully included in all parameter counts reporting.

4. Bell state preparation (zero parameters): In entangled QMARL variants, the $N$ agent qubits are initialised in a pre-entangled Bell state before the VQC layers run. A Bell state is a specific two-qubit quantum state with maximal entanglement. This state has the property that measuring one qubit \textit{instantaneously} determines the outcome of measuring the other, regardless of the spatial separation between agents.

The Bell state is prepared by a fixed two-gate circuit: a Hadamard gate ($H$) applied to the first qubit, followed by a CNOT gate targeting the second qubit:
\begin{equation}
    |00\rangle \xrightarrow{H \otimes I} 
    \frac{1}{\sqrt{2}}(|00\rangle + |10\rangle) 
    \xrightarrow{\text{CNOT}} 
    \frac{1}{\sqrt{2}}(|00\rangle + |11\rangle) = |\Phi+\rangle
\end{equation}

The Hadamard and CNOT gates are fixed, standard quantum gates with no tunable angles. The Bell state is prepared identically at the start of every episode regardless of the observation or the current policy parameters. It is a fixed quantum resource provided to the agents, not a learning component.

\paragraph{Why parameter count matters?} The standard challenge with quantum-classical comparisons is, \textit{"Is the quantum variant winning because of quantum effects, or simply because it has more parameters?"}

Table ~\ref{tab:param_counts} shows that this is not the case in our experiments. In CHSH, the classical and quantum policies use the same number of trainable parameters, so the performance gap cannot be attributed to model capacity. In CoinGame, the hybrid QMARL actors have actor parameter counts closely comparable to the classical actors, with most trainable parameters residing in the classical preprocessing layer rather than in the compact VQC core. In CoopNav, the quantum actor parameter counts remain comparable to the classical actor across the tested qubit configurations, and smaller quantum actors can still outperform the classical baseline.

Overall, the parameter analysis suggests that the observed performance differences are unlikely to be explained solely by larger model size. Instead, the results point to different mechanisms across environments: entanglement-based coordination in CHSH, VQC policy expressiveness and hybrid actor-critic structure in CoopNav and no consistent quantum benefit in CoinGame.

\newpage

\end{document}

%% file: Tables/CoopNav_Ablation_v1.0.tex
\begin{table}[h]
\centering
\caption{CoopNav full ablation: success rate across all actor/critic/qubit/depth configurations ($5{\times}5$ grid, $p_\text{slip}=0.10$, 10 seeds).
}
\label{tab:cn_ablation}
\begin{tabular}{p{2.4cm} p{1.4cm} p{1.4cm} r r r r r}
\toprule
\textbf{Hybridisation}
  & \textbf{Actor}
  & \textbf{Critic}
  & \textbf{Qubits}
  & \textbf{Depth}
  & \textbf{SR (Mean)}
  & \textbf{Std}
  & \textbf{SR (Best)} \\
\midrule
Hybrid QMARL   & Quantum   & Classical & 6          & 3          & \textbf{0.850} & 0.028 & 0.903 \\
Pure QMARL     & Quantum   & Quantum   & 6          & 4          & 0.833          & 0.025 & 0.871 \\
Hybrid QMARL   & Quantum   & Classical & 6          & 4          & 0.829          & 0.035 & 0.883 \\
Hybrid QMARL   & Quantum   & Classical & 3          & 3          & 0.814          & 0.035 & 0.871 \\
Hybrid QMARL   & Quantum   & Classical & 4          & 4          & 0.809          & 0.026 & 0.852 \\
Hybrid QMARL   & Quantum   & Classical & 4          & 3          & 0.807          & 0.029 & 0.836 \\
Hybrid QMARL   & Quantum   & Classical & 3          & 4          & 0.798          & 0.047 & 0.878 \\
Pure QMARL     & Quantum   & Quantum   & 4          & 3          & 0.698          & 0.241 & 0.863 \\
Pure QMARL     & Quantum   & Quantum   & 6          & 3          & 0.678          & 0.235 & 0.866 \\
Pure QMARL     & Quantum   & Quantum   & 3          & 4          & 0.667          & 0.233 & 0.799 \\
Hybrid QMARL   & Classical & Quantum   & $-$        & $-$        & 0.549          & 0.140 & 0.726 \\
Pure QMARL     & Quantum   & Quantum   & 4          & 4          & 0.547          & 0.323 & 0.825 \\
Pure QMARL     & Quantum   & Quantum   & 3          & 3          & 0.545          & 0.296 & 0.825 \\
\midrule
Classical MARL & Classical & Classical & $-$        & $-$        & 0.338          & 0.198 & 0.615 \\
\bottomrule
\end{tabular}
\end{table}